\documentclass[12pt,3p,times,sort,compress]{elsarticle}

\usepackage{amssymb}
\usepackage{amsmath}
\usepackage{relsize}
\usepackage{booktabs}
\usepackage{hyperref}
\usepackage[utf8]{inputenc}
\usepackage[T1]{fontenc} 
\usepackage{multirow}
\usepackage[table,xcdraw]{xcolor}
\usepackage[dvipsnames]{xcolor} 
\usepackage{placeins}
\usepackage{orcidlink}
\usepackage{bbm}
\usepackage{tabularx}
\usepackage{threeparttable}
\usepackage{array}
\usepackage{ragged2e}
\usepackage{longtable}
\usepackage{pdflscape}
\usepackage{multicol}

\long\def\dred#1{{\color{BrickRed}{#1}\color{black}}}

\long\def\dgreen#1{{\color{ForestGreen}{#1}\color{black}}}

\begin{document}

\title{Foundation models for electricity price forecasting and battery arbitrage: Can they replace market-specific forecasting models?}

\author[ONAS,ORBI]{Arkadiusz Lipiecki \orcidlink{0000-0003-1118-0388}}
\author[ORBI,CORE]{Rafał Weron \orcidlink{0000-0003-1619-5239}\corref{cor1}}
\address[ONAS]{Department of Computational Social Science, Wrocław University of Science and Technology, Poland}
\address[ORBI]{Department of Operations Research and Business Intelligence, Wrocław University of Science and Technology, Poland}
\address[CORE]{Center for Research in Energy (CoRE), Aarhus University, 8000 Aarhus C, Denmark}
\cortext[cor1]{Corresponding author; \textit{email:} rafal.weron@pwr.edu.pl}

\begin{abstract}

Foundation models promise accurate forecasts with little or no task-specific training, but whether they can replace models designed specifically for electricity price forecasting remains unclear. We compare nine variants from five foundation model families, evaluated in zero-shot mode, with two state-of-the-art electricity price forecasting benchmarks in Germany, Poland, and Spain over 2021-2025. Their performance is assessed in terms of point and probabilistic forecasting accuracy, as well as economic value in battery energy storage arbitrage. Only the TabPFN models consistently and significantly outperform the benchmarks across all three markets and all statistical measures. However, this statistical dominance does not translate directly into economic dominance: TabPFN performs best under unlimited bids and riskier quantile-based strategies, whereas the Distributional Deep Neural Network benchmark is more profitable when risk tolerance is lower. Thus, foundation models cannot universally replace market-specific models, and their value depends on both model architecture and the decision problem.
\end{abstract}


\maketitle

\begin{multicols}{2}
{\scriptsize
\tableofcontents
}
\end{multicols}

\section{Introduction}

Electricity price forecasts are a fundamental input to the decision-making processes of power companies \citep{ali:chi:zar:woo:19,bel:etal:22}. Because expectations about future outcomes shape most operational decisions \citep{mak:etal:24:M6}, even modest improvements in forecast accuracy can have substantial economic value \citep{hon:pin:etal:20,ser:wer:25}. Yet predicting electricity prices is anything but easy.

Day-ahead electricity prices exhibit pronounced daily and weekly seasonality, abrupt spikes, heavy tails, and, in many markets, negative values \citep{mac:uni:wer:23}. They depend not only on their own recent history but also on weather-driven demand, renewable generation, fuel and emission prices, plant availability, transmission constraints, and market design \citep{ghe:zie:26,mas:deb:ame:kaz:26}. Moreover, prices for all load periods of the following day are determined simultaneously using the same information set. They should therefore not be viewed as unrelated univariate time series  \citep{zie:wer:18}. Any successful forecasting approach must account for both the complex temporal dependence of electricity prices and the market-relevant information available at the forecast origin.

A further challenge concerns the form in which future prices should be represented. Point forecasts, such as conditional means or medians, are easy to interpret and can be incorporated directly into many operational procedures. However, they provide no information about forecast uncertainty and may create a misleading impression of precision \citep{pet:etal:22}. This limitation becomes particularly important when forecast errors have asymmetric consequences, extreme price realizations are costly, or decisions are subject to risk constraints \citep{chi:zam:zar:pal:18,uni:wer:21}. In such settings, probabilistic forecasts, expressed as quantiles \citep{mac:ser:uni:24} or full predictive distributions \citep{bru:mat:por:vit:19}, are more informative because they describe both the expected outcome and the uncertainty surrounding it \citep{now:wer:18}.

Instead of estimating a separate model from scratch for each market and forecasting task, a \textit{foundation model} (FM) is pretrained on a large and diverse collection of empirical datasets, purely synthetic data, or both, and can then be applied without retraining in the so-called \textit{zero-shot} mode \citep{das:etal:24}. In principle, such a model can transfer information across time series, reduce the need for manual feature engineering, and provide useful point or probabilistic forecasts even when only a short history is available. 

Foundation models have recently attracted growing attention in \textit{electricity price forecasting} (EPF). Hornek et al.~\cite{hor:etal:25:EEM} consider five European day-ahead markets and six time series FMs, but they only benchmark point forecasting performance, do not consider exogenous predictors, and limit the testing period to a single year. Marchesi et al.~\cite{mar:bal:bru:25:Energies} go beyond point forecasts and assess predictive performance in a probabilistic setting in four markets, but only for the Moirai model and a single year of testing. Lettner et al.~\cite{let:etal:26} consider probabilistic forecasting with Moirai and Chronos models, but limited to a German market, a single year of testing, and no assessment of the economic value of forecasts. Yu et al.~\cite{yu:etal:26:pricefm} introduce a FM for day-ahead electricity markets and benchmark it against three general time series FM families on a dataset comprising 24 European countries. However, the testing period is again limited to a single year and does not consider the performance of forecasts in a market-related decision problem. Ponyuenyong et al.~\cite{pon:etal:26} evaluate point forecasts from a large set of time series FMs, but only in a Singapore market and with less than half a year of testing. 

The strength of the benchmarks also varies across studies. Hornek et al.~\cite{hor:etal:25:EEM} primarily compare FMs with generic statistical methods and relatively simple machine learning models. Ponyuenyong et al.~\cite{pon:etal:26} consider more complex deep learning benchmarks based on long short-term memory and convolutional neural networks but lack comparison with EPF-specific models. By contrast, Marchesi et al.~\cite{mar:bal:bru:25:Energies} include a model developed for probabilistic EPF by Marcjasz et al.~\cite{mar:nar:wer:zie:23}.

The key question is whether broad pretraining offers an advantage over market-specific EPF models. Existing studies do not establish whether zero-shot FMs can consistently outperform such models across different day-ahead markets and over an extended test period, nor whether gains in statistical forecast accuracy translate into greater economic value. We address this gap through a common evaluation comprising (i) point forecasting, (ii) probabilistic forecasting, (iii) economic assessment of forecast accuracy, (iv) both time series and tabular foundation models, and (v) established EPF benchmarks.

\subsection{Our contribution}

This study provides a large-scale empirical comparison of foundation models for probabilistic day-ahead electricity price forecasting. We consider nine variants from five foundation-model families: Chronos-2, Chronos-2-synth, Chronos-2-small, Mitra, Moirai-2, TabPFN-2, TabPFN-3, TabPFN-TS-3, and TimesFM-2.5. These models are compared with two state-of-the-art EPF benchmarks. The empirical analysis covers a common five-year test period, from 2021 to 2025, in three structurally different European power markets: Germany, Poland, and Spain. This setting allows us to examine whether the relative performance of the models is robust across markets with different generation mixes, levels of renewable penetration, and price dynamics.

In line with recent developments in the EPF literature \citep{uni:wer:21,hir:zie:26}, we evaluate the models from both statistical and economic perspectives. Point forecast accuracy is assessed using the Mean Absolute Error (MAE) and Root Mean Squared Error (RMSE), while probabilistic performance is evaluated using the Continuous Ranked Probability Score (CRPS), a strictly proper scoring rule for probabilistic forecasts \citep{gne:kat:14,now:wer:18}. To complement the statistical evaluation, we assess the economic value of the forecasts in a Battery Energy Storage System (BESS) arbitrage application, using both quantile-based decisions and an unlimited-bid setting \citep{mar:nar:wer:zie:23,hir:zie:26}.

The study yields two main findings. First, pretrained foundation models with native support for exogenous variables perform competitively against market-specific EPF benchmarks, but only the TabPFN family achieves superior statistical accuracy across all considered markets. Second, the economic evaluation provides a more nuanced picture: TabPFN dominates under riskier battery trading strategies, whereas the Distributional Deep Neural Network (DDNN) benchmark is the most profitable model when risk tolerance is lower.

The remainder of the paper is structured as follows. Section~\ref{sec:Literature} reviews the emerging literature on foundation models. Section~\ref{sec:Datasets} briefly describes the datasets. Section~\ref{sec:FM:Def} then introduces the time series and tabular foundation models and explains their input structures and forecasting setups, while Section~\ref{sec:Benchmark:Def} presents the two benchmarks. Section~\ref{sec:Empirical:Evidence} presents the statistical and economic evaluation. Finally, Section~\ref{sec:Conclusions} concludes.

\section{Literature review}
\label{sec:Literature}

According to Maciejowska et al.~\cite{mac:uni:wer:23}, recent EPF research is characterized by three broad trends: a shift toward high-dimensional statistical and machine learning methods, increasing interest in probabilistic forecasts, and growing emphasis on economic evaluation. The literature review below focuses on the first trend and the emerging use of foundation models, whereas our empirical study addresses all three.

\subsection{Point forecasting with foundation models}

Hornek et al.~\cite{hor:etal:25:EEM} compare six models (Chronos-Bolt, Chronos-T5, TimesFM, Moirai, Time-MoE, and TimeGPT) in a point forecasting setup. The test period spans one year (2024) for each of the five European day-ahead markets: Austria, Belgium, France, Germany, and the Netherlands. Chronos-Bolt and Time-MoE are the strongest FMs, but none significantly outperforms MSTL, a comparatively simple multiple-seasonal decomposition benchmark \citep{ban:hyn:ber:25}. The authors conclude that zero-shot forecasting is feasible, but pretraining alone does not guarantee superior accuracy.

More encouraging results are reported in a point forecasting study of the real-time Singapore market, which operates with half-hourly load periods. Ponyuenyong et al.~\cite{pon:etal:26} combine Tiny Time Mixers, Moirai, MOMENT, and TimesFM (zero-shot and fine-tuned versions) with a regularization strategy designed to improve performance during price spikes. The foundation models outperform the ARIMA, LSTM, and CNN-LSTM benchmarks by up to 37.4\% in the Mean Absolute Percentage Error (MAPE) over a test set comprising 10\% of the 2021-2024 sample; note that MAPE should actually be avoided in EPF \citep{wer:14,lag:mar:sch:wer:21}. The authors argue that FMs may be particularly useful when the price process is highly nonlinear and conventional models struggle to accommodate extreme observations. However, the study considers generic benchmarks rather than EPF-specific models such as those discussed in Sec.~\ref{sec:Benchmark:Def} and uses real-time market data. Therefore, its conclusions may not be directly transferable to day-ahead markets.

Sayghe et al.~\cite{say:etal:26} propose GridFM, a physics-informed foundation model for the joint point forecasting of electricity load, locational marginal prices, carbon emissions, and renewable generation using 11 years of real-time data from the New York Independent System Operator (NYISO) at five-minute resolution. The test set covers the final two years, 2023-2024, and the forecasting horizons range from 1 to 24 hours. The model combines a foundation model with a frequency-domain adaptation layer, graph-based representations of grid topology, physical constraints, multitask learning, and explainability tools. The study focuses more broadly on real-time, multitask power-system forecasting rather than specifically on EPF. Nevertheless, for the EPF task, it reports a 15.9 to 23.2\% reduction in MAPE relative to the baseline foundation models TimesFM, Chronos, and Moirai-MoE. The study reports CRPS and coverage statistics in Sec.~5.9 but does not discuss them in detail. As with \cite{pon:etal:26}, its conclusions may not be directly transferable to day-ahead markets.

\subsection{Probabilistic forecasting with foundation models}

Marchesi et al.~\cite{mar:bal:bru:25:Energies} study Moirai in zero-shot and fine-tuned probabilistic forecasting settings. Considering day-ahead data from four European bidding zones (Belgium, Germany, Spain, and Sweden SE$_3$) and test periods spanning a full year (1.10.2023 to 30.09.2024), they find that zero-shot forecasts are promising, but generally less accurate in terms of the CRPS than those of the DDNN benchmark; see Sec.~\ref{sec:Benchmark:Def} or \cite{mar:nar:wer:zie:23}. Fine-tuning improves calibration, whereas the gains from including exogenous variables are limited. The latter result points to one of the major weaknesses of many FMs -- if they were designed primarily for univariate forecasting, they may not fully exploit the covariates. 
This is a serious limitation in EPF. At the time day-ahead bids are submitted, market participants typically have access to forecasts of load and renewable generation. Ignoring this information puts even a sophisticated pretrained model at a disadvantage relative to a well-specified regression or neural network. 

Two recent approaches address this issue more directly. PriceFM is a domain-specific foundation model trained on data from 38 interconnected European bidding zones \citep{yu:etal:26:pricefm}. It incorporates load, wind, and solar forecasts, as well as information on transmission topology, within a probabilistic EPF framework. The reported Average Quantile Loss (AQL; similar to the CRPS) for 2025, together with the MAE and RMSE, indicates strong zero-shot and fine-tuned performance. The improvements are substantial relative to naive benchmarks and more moderate relative to FM baselines (Chronos, Moirai, TimeMoE, and TimesFM) and models specialized in spatial forecasting (including PatchTST and GraphDiffusion). Nevertheless, Ref.~\cite{yu:etal:26:pricefm} points to a promising direction for future research: combining large-scale pretraining with spatial structures and covariates that reflect the operation of electricity markets. However, the PriceFM model includes recent EPF data (with 2024 being the last training year) in its training corpus, which would not allow us to consider an extensive five-year testing period.

ApolloPFN takes a different approach \cite{pot:etal:26:apollo}. It is a time-aware prior-fitted network trained on synthetically generated data and designed to incorporate future exogenous variables directly. Across the same five datasets as in \cite{lag:mar:sch:wer:21}, it performs on par with TabPFN-TS-2 in terms of the CRPS and considerably better than Chronos-large, Moirai-large, Sundial-base, and classical time series benchmarks such as AutoARIMAX and Prophet. Although the reported performance is very promising, ApolloPFN is a proprietary model and is therefore not included in our study. 

Finally, Lettner et al.~\cite{let:etal:26} compare Moirai-1 and ChronosX with two task-specific probabilistic EPF models, NHITS combined with Quantile Regression Averaging (QRA) and a Transformer-conditioned Normalizing Flow, on the German-Luxembourg bidding zone day-ahead prices in 2024. Fine-tuned foundation models achieve the best CRPS, Energy Score, and calibration, while their zero-shot variants also outperform simple empirical benchmarks. However, a carefully configured NHITS+QRA model performs relatively close to the FMs and can benefit more consistently from additional exogenous and cross-market information. The authors conclude that the modest predictive gains offered by FMs should be weighed against their substantially greater computational requirements.

\subsection{Components of larger forecasting systems}

Foundation models can also be used as components of larger forecasting systems. FutureBoosting uses a pretrained time series FM to predict explanatory variables that are unavailable at the forecast origin and passes these predictions to a regression model~\citep{qui:etal:26}. This hybrid approach outperforms both standalone FMs and conventional regression benchmarks in the reported point forecasting experiments. The takeaway is that a foundation model can generate covariates that improve a simpler downstream model.

A related lesson follows from Temporal Hierarchy Forecasting (THieF). Lipiecki et al.~\cite{lip:bil:kou:wer:26} show that reconciling forecasts for hourly, multi-hour block, and baseload products improves the accuracy of linear regression, neural network, gradient-boosted tree, and Mitra forecasts. This suggests that the information contained in the market structure and in relationships across temporal aggregation levels can be at least as valuable as increasing model complexity. A foundation model should therefore be viewed as one element of the forecasting pipeline, not as a substitute for careful problem formulation, data preprocessing, or forecast combination.

\subsection{Conclusions}

Overall, the existing literature has separately examined point and probabilistic zero-shot forecasting, the use of exogenous information, and hybrid FM-based forecasting systems. However, the evidence is based predominantly on short test periods, and no study reviewed above jointly evaluates FMs against strong EPF-specific benchmarks in terms of point forecasting accuracy, probabilistic forecasting accuracy, and economic value over a long test period. The empirical analysis below is designed to address this gap.

\begin{figure}[tb]
    \centering
    \includegraphics[width=.45\linewidth]{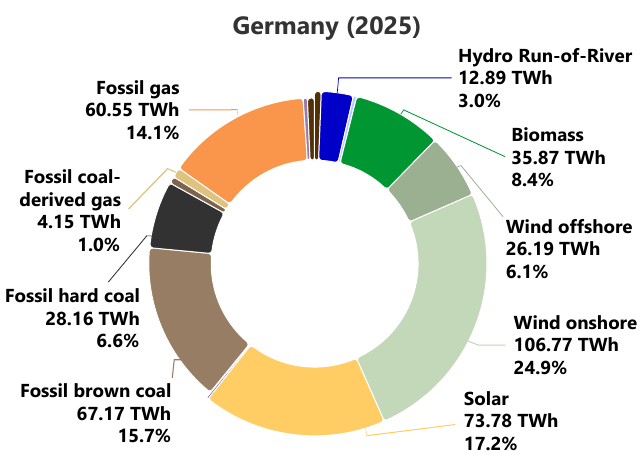}~~~~~~
    \includegraphics[width=.45\linewidth]{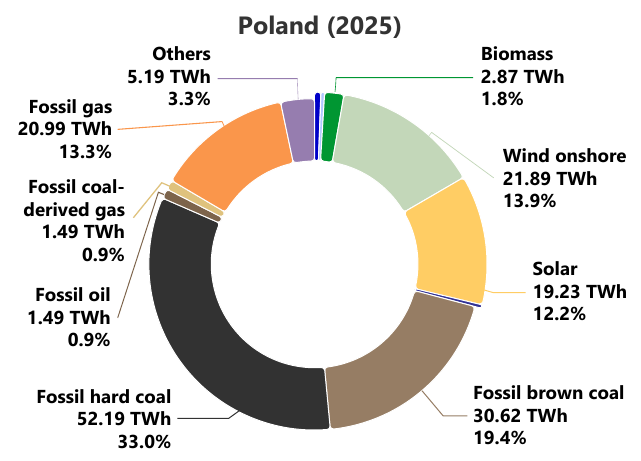}\\[6pt]
    \includegraphics[width=.45\linewidth]{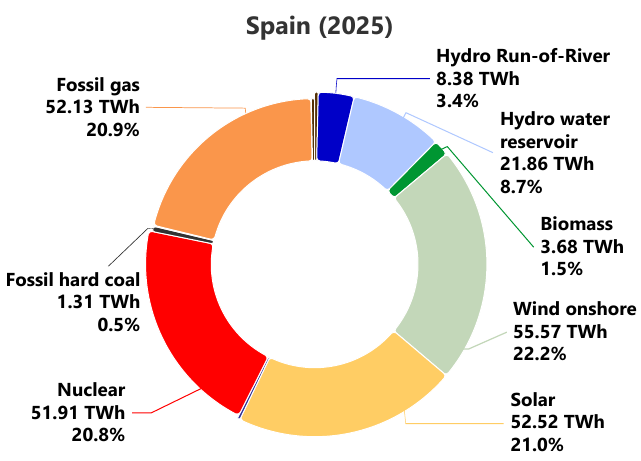}~~~~~~
    \includegraphics[width=.45\linewidth]{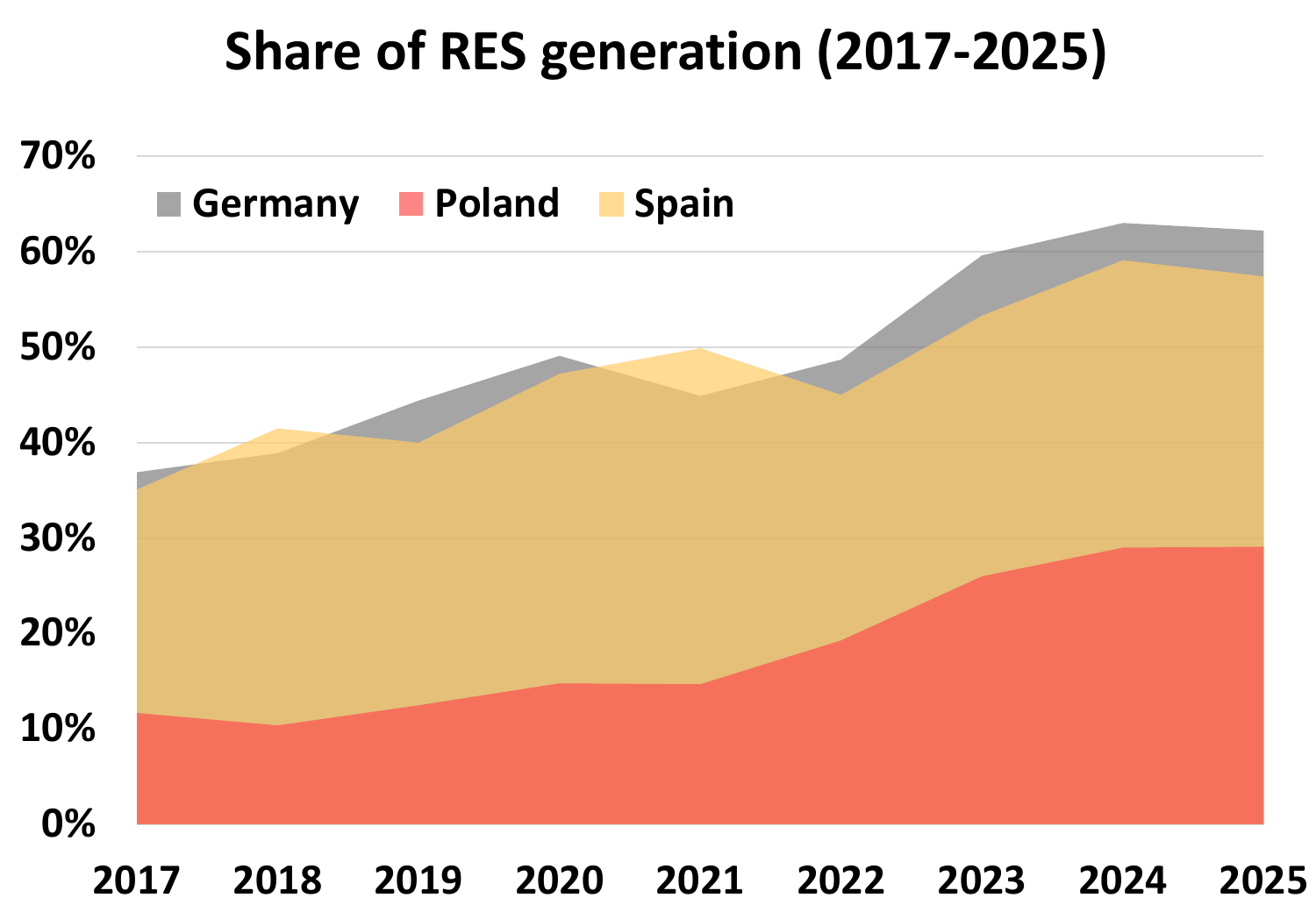}
    \caption{Public net electricity generation in Germany, Poland, and Spain in 2025, together with the share of renewable energy sources (RES) in electricity generation in the three countries over the 2017-2025 study period.
    \textit{Data source:} \url{https://www.energy-charts.info}.}
    \label{fig:Generation:DE:PL:ES}
\end{figure}

\begin{figure*}[p]
    \centering
    \includegraphics[width=.85\linewidth]{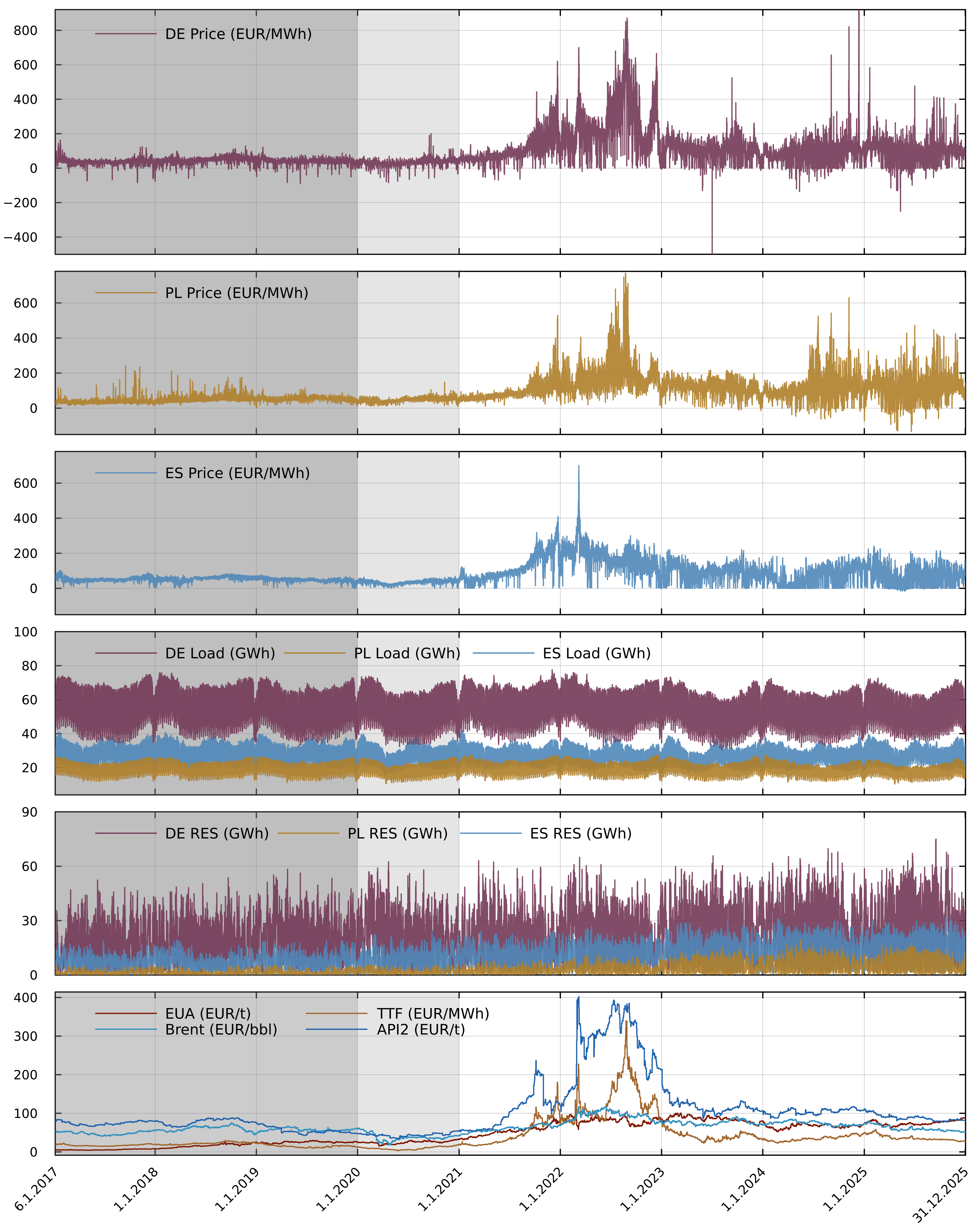}
    \caption{\textit{Top three panels:} German (DE), Polish (PL), and Spanish (ES) day-ahead electricity prices from 6 January 2017 to 31 December 2025. \textit{Two middle panels:} Day-ahead load and renewable generation (RES) forecasts in the three countries. \textit{Bottom panel:} Four fundamental variables common for all three markets: European Union Allowance (EUA) prices, natural gas (TTF) prices, crude oil (Brent) prices, and coal (API2) prices. The dark shaded area indicates the initial training window for the point-forecasting models, that is, the data used to estimate LEAR and Mitra when generating forecasts for the first day of the initial postprocessing window, shown by the light shaded area. Taken together, the dark and light shaded areas indicate the initial training window for the models that directly produce probabilistic forecasts for the first day of the test period, 1 January 2021.}
    \label{fig:datasets}
\end{figure*}

\section{Datasets}
\label{sec:Datasets}

The data used in this study are publicly available and can be downloaded from ENTSO-E (\url{https://transparency.entsoe.eu}; day-ahead prices, day-ahead load forecasts, and day-ahead onshore wind, offshore wind, and solar generation forecasts) and Investing.com (\url{https://www.investing.com/}; closing prices of carbon emission allowances, natural gas, crude oil, and coal futures). The datasets span the period from 6 January 2017 to 31 December 2025, with a five-year out-of-sample test period beginning on 1 January 2021. 

We consider three major European markets -- Germany, Poland, and Spain -- with diverse characteristics and levels of renewable energy sources (RES) penetration. This variation provides a useful test of whether model performance is robust across systems with different generation mixes and price dynamics. Figure~\ref{fig:Generation:DE:PL:ES} illustrates these structural differences by showing the public net electricity generation mix in 2025 and the evolution of the share of RES in electricity generation over the 2017-2025 study period. The figure highlights the comparatively high shares of wind and solar generation in Germany and Spain, the stronger role of coal in Poland, and the presence of nuclear generation in Spain, as well as the increasing contribution of RES over time.

Day-ahead prices, see the top three panels in Figure~\ref{fig:datasets}, represent the following bidding zones: BZN$|$DE-LU for Germany, BZN$|$PL for Poland, and BZN$|$ES for Spain. The day-ahead load and renewable generation forecasts correspond to the countries -- Germany (DE), Poland (PL), and Spain (ES) -- rather than to the bidding zones. As in \cite{mar:nar:wer:zie:23,uni:26:VST}, the \textit{renewable generation} forecast is calculated as the sum of the day-ahead onshore wind, offshore wind (only for Germany), and solar generation forecasts. Because some ENTSO-E series are available at a 15-minute resolution, we aggregate them to hourly values. 

As in \cite{mar:nar:wer:zie:23,uni:zie:26}, we consider four fundamental variables: European Union Allowance (EUA) prices, natural gas prices from the Title Transfer Facility (TTF) virtual trading point in the Netherlands, Brent crude oil prices, and API2 coal prices. They are represented by the most recent closing prices of nearest-to-delivery futures contracts (yearly for EUA, monthly for others) available at the time of bidding in the day-ahead power market, corresponding to a two-day lag relative to the target day, see the bottom panel in Figure~\ref{fig:datasets}. 

All time series were preprocessed to account for transitions to and from daylight saving time (DST). Missing values occurring during the transition to DST were replaced with observations from the preceding hour. Duplicate values occurring during the transition from DST were replaced with their arithmetic mean. Any missing values in the day-ahead price and load forecasts were replaced with the value from the preceding hour or, if that value was also missing, with the value from the preceding day. Missing values in onshore wind, offshore wind, and solar generation forecasts were replaced with zeros.

\section{Foundation models}
\label{sec:FM:Def}

Recent years have witnessed rapid advances in foundation models for both time series and tabular prediction tasks. Although these two applications involve distinct challenges and data structures, the FMs considered in this study share several features. All are based on the Transformer architecture \citep{vas:etal:17}, which comprises encoder and decoder stacks that rely on attention mechanisms but process information differently. The encoder has access to the entire input sequence and transforms it into a latent representation. The decoder uses masked attention, restricting information flow so that the output associated with each sequence element depends only on preceding elements, thus imposing an autoregressive structure on the output. Modern architectures often use just one of these components, called encoder-only, first implemented in BERT, and decoder-only, popularized by GPT-1~\citep{min:eta:23:LLM_rev}.

In what follows, we describe the FMs used in this study, beginning with the time series models and then turning to the tabular models. We focus on their \textit{zero-shot} performance; that is, their ability to generate forecasts for previously unseen series without task-specific weight updates, relying instead on the in-context learning paradigm~\citep{hol:etal:25}. 

\begin{table}[tb]
\centering
\caption{Foundation models and benchmarks considered in this study. Except for the LEAR benchmark, model size is reported in millions (M) of parameters. Pretraining data are classified as \textit{synthetic} or \textit{mixed}. MLP, PDF, and TS stand for multilayer perceptron, probability density function, and time series, respectively, while $\rightarrow$ indicates the model output; see the main text for details. Models marked with $\dagger$ are not licensed for commercial use free of charge.}
\label{tab:FMs}
\footnotesize
\begin{tabular}{lrlrll}
\toprule
\textbf{Model} & \textbf{Release date} & \textbf{Type} & \textbf{Size} & \textbf{Pretraining} & \textbf{Architecture} \\
\midrule
\multicolumn{6}{c}{\textit{Foundation models}} \\[3pt]
Chronos-2       & {Oct. 2025}~\citep{ans:etal:25} & \multirow{3}{*}{$\left.\rule{0pt}{18pt}\right\}$ TS} & 120M  & Mixed     & \multirow{3}{*}{$\left.\rule{0pt}{18pt}\right\}$ Encoder-only $\rightarrow$ Quantiles} \\
Chronos-2-synth & {Nov. 2025} & & 120M  & Synthetic &  \\
Chronos-2-small & {Nov. 2025}  & & 28M   & Mixed     &  \\[3pt]
Moirai-2${^\dagger}$        & Aug. 2025~\citep{liu:etal:26} & TS & 11M & Mixed     & Decoder-only $\rightarrow$ Quantiles \\[3pt]
TimesFM-2.5     & Sep. 2025~\citep{das:etal:24} & TS & 200M  & Mixed     & Decoder-only $\rightarrow$ Quantiles \\[3pt]
TabPFN-2        & Jan. 2025~\citep{hol:etal:25} & Tabular     & 11M   & \multirow{3}{*}{$\left.\rule{0pt}{18pt}\right\}$ Synthetic} & \multirow{2}{*}{$\left.\rule{0pt}{12pt}\right\}$ Encoder-only $\rightarrow$ Discretized PDF} \\
TabPFN-3${^\dagger}$        & May 2026~\citep{gri:etal:26}  & Tabular     & 58M   & &  \\
TabPFN-TS-3${^\dagger}$     & May 2026~\citep{hoo:etal:26}  & TS+Tabular & 58M   & & TS feature generation $\rightarrow$ TabPFN-3 \\[3pt]
Mitra           & Jul. 2025~\citep{zha:etal:25} & Tabular     & 72M   & Synthetic & Encoder-only $\rightarrow$ Mean \\
\midrule
\multicolumn{6}{c}{\textit{Benchmarks}} \\[3pt]
LEAR & Jul. 2021 \cite{lag:mar:sch:wer:21} & Tabular & 250   &  & Linear $\rightarrow$ Mean \\[3pt]
DDNN-JSU & Sep. 2023 \cite{mar:nar:wer:zie:23} & Tabular  & 0.4M  &  & MLP $\rightarrow$ PDF parameters \\
\bottomrule
\end{tabular}
\end{table}

\subsection{Time series models}
\label{ssec:TSFM}

\subsubsection{Chronos-2}
\label{sssec:Chronos}

Chronos-2 is a time series foundation model developed by Amazon for general-purpose forecasting tasks~\citep{ans:etal:25}. Released in October 2025, it uses an encoder-only Transformer architecture with time-attention and group-attention layers. These layers are followed by a quantile head that produces forecasts at 21 probability levels: 19 equally spaced levels between 0.05 and 0.95, together with two additional tail levels, 0.01 and 0.99, intended to improve the representation of the predictive distribution tails and allow the retrieval of percentile forecasts by interpolation. Note that Chronos-2 is the only FM in this study that natively supports multivariate forecasting.

We consider three variants: the default \textbf{Chronos-2} model with 120 million parameters, \textbf{Chronos-2-synth}, which is pretrained exclusively on synthetic data, and \textbf{Chronos-2-small}, which contains 28 million parameters and is approximately one-quarter the size of the default model.

\subsubsection{Moirai-2}
\label{sssec:Moirai}

\textbf{Moirai-2} is a decoder-only Transformer developed by Salesforce AI Research and released in August 2025~\citep{liu:etal:26}. Compared with its predecessors, Moirai-1 and Moirai-MoE~\citep{liu:etal:25}, it replaces mixture-distribution modeling with direct quantile estimation, similarly to Chronos-2. Although the quantile head of Moirai-2 is limited to nine deciles, the model implements an autoregressive multi-quantile decoding algorithm that increases the resolution of the predictive distribution. It does so by feeding the predicted deciles back into the model, generating additional sets of deciles, and then sampling quantiles at the required probability levels.

With 11.4 million parameters, Moirai-2 is one of the smallest FMs considered in this study. It is also the only model from the Moirai family that does not support covariates. However, in our initial tests, both Moirai-1 and Moirai-MoE performed considerably worse than Moirai-2, so we include only the latter in the final comparison.

\subsubsection{TimesFM-2.5}
\label{sssec:TimesFM}

Developed by Google Research and released in September 2025, \textbf{TimesFM-2.5} is a time series FM that, like Moirai-2, uses a decoder-only architecture~\citep{das:etal:24}. With approximately 200 million parameters, it is the largest model considered in this study. To generate probabilistic forecasts, TimesFM-2.5 uses a quantile head that directly produces forecasts for nine decile levels. Although the model does not natively incorporate covariates, they can be included through an external regression mechanism known as XReg. We use the default XReg pipeline, in which the residuals of the TimesFM forecasts over the context window are regressed linearly on the supplied covariates. The final forecast is obtained by adding the resulting regression component to the original TimesFM output.

\subsection{Tabular models}
\label{ssec:TabularFM}

\subsubsection{TabPFN}
\label{sssec:TabPFN}

TabPFN is a tabular foundation model developed by Prior Labs for classification and regression tasks~\citep{hol:etal:25}. It is a Transformer pretrained solely on synthetic datasets according to the Prior-Fitted Network (PFN) learning paradigm. 
The central idea is to sample datasets from a prior distribution over generative processes and train the model on these synthetic datasets in a supervised-learning framework.

PFNs use \textit{in-context learning}, i.e., the training data are supplied to the model together with the predictor values corresponding to the unknown test target, allowing it to generate a predictive distribution for that target. As a result, the model can infer the relationship between predictors and responses from the supplied data and produce predictions for unseen observations in a single forward pass, without task-specific training.

In this study, we evaluate three variants of TabPFN:
\begin{itemize}\itemsep0em
    \item \textbf{TabPFN-2} is a lightweight model with 11 million parameters, released in January 2025. Its architecture consists of a linear encoder followed by twelve TabPFN layers containing column-wise attention, row-wise attention, and a multilayer perceptron~\citep{hol:etal:25}.
    \item \textbf{TabPFN-3}, a substantially larger model with 58 million parameters that can process up to 100 times more observations or 40 times more features~\citep{gri:etal:26}. It replaces separate row-wise and column-wise attention with full-row embeddings. This version of TabPFN was released in May 2026.
    \item \textbf{TabPFN-TS-3}, a data-processing pipeline that performs feature generation and augmentation for time series data, transforms the forecasting task into a tabular regression problem to be solved by TabPFN~\citep{hoo:etal:26}. We use it in tandem with TabPFN-3 checkpoint dedicated to time series regression, also released in May 2026.
\end{itemize}
For all TabPFN regression models, the output layer represents the predictive distribution over a fixed set of bins spanning the target space. Quantile forecasts are then derived from this discretized distribution.

\subsubsection{Mitra with conformal prediction}
\label{sssec:Mitra}

Mitra is a tabular FM released in July 2025 by Amazon~\citep{zha:etal:25}. It builds directly on the PFN framework but places greater emphasis on the choice of priors used during pretraining. In addition to the structural causal models employed by TabPFN~\citep{hol:etal:25}, Mitra uses priors based on a mixture of tree-based models, including decision trees, extremely randomized trees, gradient boosting, and random forests. These models are sampled either directly or indirectly after being fitted to synthetic data.

Architecturally, Mitra is a 12-layer encoder-only Transformer, with each layer comprising row-wise attention, column-wise attention, and a multilayer perceptron, similarly to TabPFN-2. Because the regression variant of Mitra does not produce probabilistic forecasts directly, we postprocess its point forecasts using the same procedure as in the case of LEAR+CP. The resulting model is denoted by \textbf{Mitra+CP}.

\subsection{Context data}
\label{ssec:Features}

For time series FMs that incorporate exogenous predictors, namely the three Chronos-2 variants, TimesFM-2.5, and TabPFN-TS-3, the relevant inputs are provided as complete time series without a predefined lag structure. For the \textbf{Chronos-2} models, which support multivariate forecasting, the following is used as past context data $(\mathcal{X}_d, \mathcal{Y}_d)$:
\begin{align}
\label{eqn:train_mvts}
    \mathcal{X}_d =~ & \Big\{ \left(\boldsymbol{\hat{L}}_{k}\right)_{k=d-D}^{d-1},\;
    \left(\boldsymbol{\hat{R}}_{k}\right)_{k=d-D}^{d-1},\;  
    \left(\text{EUA}_{k-2}\right)_{k=d-D}^{d-1},\; 
    \nonumber \\
    & \left(\text{TTF}_{k-2}\right)_{k=d-D}^{d-1}, \left(\text{Brent}_{k-2}\right)_{k=d-D}^{d-1},\; \left(\text{API2}_{k-2}\right)_{k=d-D}^{d-1} \Big\}, \nonumber \\
    \mathcal{Y}_D =~ & \left(\boldsymbol{p}_{k}\right)_{k=d-D}^{d-1},
\end{align}
where $\boldsymbol{p}_{k}=(p_{k, 1}, ..., p_{k,24})$ is a vector of 24 hourly prices for day $k$, $\boldsymbol{\hat{L}}_{k}=(\hat{L}_{k, 1}, ..., \hat{L}_{k,24})$ is a vector of 24 hourly day-ahead load forecasts for day $k$, $\boldsymbol{\hat{R}}_{k}=(\hat{R}_{k, 1}, ..., \hat{R}_{k,24})$ is a vector of 24 hourly day-ahead renewable generation forecasts for day $k$, and $D=1449$ is the length of past context data. Note that we define renewable generation as the sum of onshore wind, offshore wind, and solar generation, see Section~\ref{sec:Datasets}.
Additionally, we include four market variables: European carbon allowances (EUA$_{d-2}$), natural gas (TTF$_{d-2}$), crude oil (Brent$_{d-2}$), and coal (API2$_{d-2}$). All four correspond to the most recent closing futures prices available on day $d-2$, see Section~\ref{sec:Datasets} for details.
Then, at the prediction stage, the following future context data is provided:
\begin{align}
\label{eqn:future_mvts}
        {X}_d = \Big\{ \boldsymbol{\hat{L}}_{d}, \boldsymbol{\hat{R}}_{d},  \text{EUA}_{d-2},  \text{TTF}_{d-2},  \text{Brent}_{d-2},  \text{API2}_{d-2} \Big\}.
\end{align}

For \textbf{TimesFM-2.5} and \textbf{TabPFN-TS-3}, the forecasting task is decomposed into 24 univariate tasks, one for each hour of the day. In this setting, the information set $(\mathcal{X}_d, \mathcal{Y}_d)$ provided as context before forecasting $p_{d, h}$ takes the form:
\begin{align}
\label{eqn:train_ts}
    \mathcal{X}_d =~ & \Big\{\left(\boldsymbol{p}^{\setminus h}_{k-1}\right)_{k=d-D}^{d-1},\; \left(\boldsymbol{\hat{L}}_{k}\right)_{k=d-D}^{d-1},\; \left(\boldsymbol{\hat{R}}_{k}\right)_{k=d-D}^{d-1},\; \left(\text{EUA}_{k-2}\right)_{k=d-D}^{d-1},\; \nonumber \\
    &  \left(\text{TTF}_{k-2}\right)_{k=d-D}^{d-1},\; \left(\text{Brent}_{k-2}\right)_{k=d-D}^{d-1},\:
    \left(\text{API2}_{k-2}\right)_{k=d-D}^{d-1} \Big\}, \nonumber \\
    \mathcal{Y}_d =~ & \left(p_{k, h}\right)_{k=d-D}^{d-1},
\end{align}
where $\boldsymbol{p}^{\setminus h}_{k}$ denotes the vector of prices for all hours except hour $h$, and the remaining notation is the same as in Eq.~\ref{eqn:train_mvts}. Analogously, the future context data $X_d$ for univariate models is the following: 
\begin{align}
\label{eqn:future_ts}
    {X}_d =~ & \Big\{ \boldsymbol{p}^{\setminus h}_{d-1},  \boldsymbol{\hat{L}}_{d},  \boldsymbol{\hat{R}}_{d},  \text{EUA}_{d-2}, \text{TTF}_{d-2},  \text{Brent}_{d-2},  \text{API2}_{d-2} \Big\}.
\end{align}
Note that, compared with the multivariate setting in Eqs.~\eqref{eqn:train_mvts} and \eqref{eqn:future_mvts}, the univariate models additionally use the one-day-lagged prices of the remaining hours $\boldsymbol{p}^{\setminus h}_{d-1}$ as exogenous variables.

Because \textbf{Moirai-2} supports neither exogenous variables nor native multivariate forecasting, we adopt a different strategy that performed better for this model in our preliminary tests. Electricity prices are treated as a single univariate series at hourly resolution, and the 24 prices for the following day are generated as a multi-horizon forecast, following the approach used in \cite{mar:bal:bru:25:Energies} for Moirai-MoE. This setup preserves cross-hour information through the common hourly price sequence while relying on a univariate forecasting model. The context length for the model was set to 8,736, corresponding to 364 days of hourly prices.

The remaining models, namely \textbf{TabPFN-2}, \textbf{TabPFN-3}, and \textbf{Mitra+CP}, are tabular regression models, hence we explicitly include lagged variables in the past context set $(\mathcal{X}_d, \mathcal{Y}_d)$. The regressors provided in $X_d$ include prices for lags $k=1,2,3,7$, day-ahead load $\boldsymbol{\hat{L}}_{d-k}$ and renewable generation forecasts $\boldsymbol{\hat{R}}_{d-k}$ for lags $k=0,1,7$, day of the week categorical variable $\text{DoW}_{d}\in\{1, ..., 7\}$ capturing weekly seasonality, and the market variables for lag 2. For day $d$, the goal is to approximate the price, its distribution, or quantiles, by: 
\begin{align}\label{eq:tabular}
    f\Big(&  \boldsymbol{p}_{d-1}, \boldsymbol{p}_{d-2}, \boldsymbol{p}_{d-3}, \boldsymbol{p}_{d-7}, \boldsymbol{\hat{L}}_{d}, \boldsymbol{\hat{L}}_{d-1}, \boldsymbol{\hat{L}}_{d-7}, \boldsymbol{\hat{R}}_{d}, \boldsymbol{\hat{R}}_{d-1}, \boldsymbol{\hat{R}}_{d-7}, \nonumber \\ 
    & \text{EUA}_{d-2}, \text{TTF}_{d-2}, \text{Brent}_{d-2}, \text{API2}_{d-2}, \text{DoW}_{d} \Big).
\end{align}
TabPFN-2 and TabPFN-3 are provided the context set of the 1,449 most observations, while Mitra uses a shorter context of 1,085 observations, because 364 out-of-sample forecasts are reserved for postprocessing.

\section{Benchmark models}
\label{sec:Benchmark:Def}

Over the last decade, regularized regression models have emerged as state-of-the-art benchmarks that are relatively straightforward to interpret and communicate to decision makers. In particular, the Lasso-Estimated AutoRegressive (LEAR) model of Lago et al.~\citep{lag:mar:sch:wer:21} performs particularly well when forecasts obtained using training windows of different lengths are averaged, an approach introduced to EPF by Hubicka et al.~\cite{hub:mar:wer:19}. Multilayer perceptrons with two, or more, hidden layers and the same set of inputs, often referred to as Deep Neural Networks \citep[DNNs;][]{lag:rid:sch:18,bru:mat:por:vit:19,lag:mar:sch:wer:21}, can outperform regularized regression models, but only when they are carefully designed, trained, and, where appropriate, combined \citep{tsc:pie:pla:rob:22,zam:ioa:zar:25}.

Both LEAR and DNN are point-forecasting models, so their outputs must be \textit{postprocessed} to obtain probabilistic forecasts \citep{van:etal:21:postprocessing,lip:uni:wer:24}. Several postprocessing schemes are available, including historical simulation, conformal prediction, isotonic distributional regression, and quantile regression averaging \citep{lip:wer:25}.

A viable alternative to postprocessing point forecasts or their associated errors is to modify the DNN so that its output layer returns the parameters of an assumed probability distribution rather than hourly price forecasts. The resulting model, termed the Distributional Deep Neural Network \citep[DDNN;][]{mar:nar:wer:zie:23}, produces not only well-calibrated and sharp probabilistic forecasts, but also point forecasts whose accuracy is comparable to that of forecasts generated by the standard DNN.

\subsection{LEAR with conformal prediction}

The first benchmark is the LEAR model of Lago et al.~\citep{lag:mar:sch:wer:21}, implemented with minor modifications. More specifically, for each hour of the following day, four linear regression models are estimated using the Lasso with two short rolling training windows of 56 and 84 days, corresponding to 8 and 12 weeks, and two long windows of 728 and 1,092 days, corresponding to 2 and 3 years. The four resulting forecasts are then averaged. Because only three years of data are available for model estimation, with the fourth year reserved for postprocessing, our implementation differs from the original specification, which used three- and four-year training windows.

Following the recommendations of Uniejewski~\cite{uni:24:reg}, the Lasso regularization parameter $\lambda$ is selected by seven-fold cross-validation at each retraining step. By contrast, the original implementation used the faster but less accurate Lasso based on the least-angle regression (LARS) algorithm. As in \citep{lag:mar:sch:wer:21}, we apply the area hyperbolic sine transformation to stabilize the variance \citep{uni:26:VST}.

LEAR solves the tabular regression task defined in Eq.~\eqref{eq:tabular} and produces point forecasts corresponding to the conditional mean. We therefore use a pipeline analogous to that employed for Mitra. LEAR is trained on a past context set comprising the 1,085 most recent observations. Unlike for the tabular FMs, the day-of-week variable $\text{DoW}_{d}$ is one-hot encoded, i.e., LEAR receives the day of the week as a set of binary dummy variables rather than as a single categorical variable.
We then apply conformal prediction~\citep[CP;][]{lip:wer:25} using a one-year rolling calibration window of out-of-sample forecast errors to obtain probabilistic forecasts. The resulting model is referred to as \textbf{LEAR+CP}.

\begin{table*}[tb]
\centering
\caption{Hyperparameter search space for the DDNN-JSU model. The number of neurons, \texttt{neurons\_i}, and the activation function, \texttt{activation\_i}, are optimized separately for each of the two hidden layers. The strength of the L1 penalty is also estimated separately for each hidden layer, \texttt{l1\_hidden\_i}, and each of the four distributional heads, \texttt{l1\_head\_i}. For dropout, \texttt{l1\_hidden\_i}, and \texttt{l1\_head\_i}, an additional binary tuning parameter determines whether the corresponding form of regularization is applied.}
\label{tab:hyperparameters}
\footnotesize
\begin{tabular}{llll}
\toprule
\textbf{Hyperparameter} & \textbf{Type} & \textbf{Range [...] / Set \{...\}} & \textbf{Sampling} \\
\midrule
\texttt{neurons\_i} & Integer & $[16, 512]$ & Uniform (per hidden layer $i$) \\
\texttt{activation\_i} & Categorical & \{\texttt{elu}, \texttt{tanh}, \texttt{relu}, \texttt{softplus}, \texttt{sigmoid}\} & Uniform (per hidden layer $i$) \\
\texttt{learning\_rate} & Float & $[10^{-5}, 10^{-1}]$ & Log-uniform \\
\texttt{dropout} & Float & $[0.0, 1.0]$ & Uniform \\
\texttt{l1\_hidden\_i} & Float & $[10^{-5}, 10^{-1}]$ & Log-uniform (per hidden layer $i$) \\
\texttt{l1\_head\_i} & Float & $[10^{-5}, 10^{-1}]$ & Log-uniform (per head layer $i$) \\
\bottomrule
\end{tabular}
\end{table*}

\subsection{DDNN-JSU}

The second benchmark, denoted \textbf{DDNN-JSU}, is a feedforward neural network with two hidden layers and an output layer that returns the parameters of a Johnson’s SU distribution for each of the 24 hours of the following day~\citep{mar:nar:wer:zie:23}. The hyperparameters are selected at the beginning of each calendar year using the preceding year as the evaluation set. The hyperparameter ranges are reported in Table~\ref{tab:hyperparameters}, and the optimization is performed using the Tree-Structured Parzen Estimator in Optuna~\citep{aki:etal:19:optuna}.

We run four independent hyperparameter-tuning studies with 500 trials each. Within each trial, the weights are reestimated every 28 days using an expanding-window approach, with the first training window consisting of 1085 data points. As a result, four independently selected hyperparameter sets are obtained for each calendar year. The networks are trained using the AdamW optimizer~\citep{los:hut:19} for a maximum of 1000 epochs, with 20\% of the training set held out for validation. To avoid overfitting, we implement early stopping with a patience of 50 epochs and a scheduler that reduces the learning rate by a factor of 10 when the validation loss does not improve for 15 consecutive epochs. Note that Marcjasz et al.~\cite{mar:nar:wer:zie:23} used a slightly different setup: 1000 hyperparameter-tuning trials, variable selection during hyperparameter optimization, the Adam optimizer, no learning-rate scheduler, an upper limit of 1024 neurons (rather than 512), and batch normalization. Our setup is computationally faster while retaining comparable predictive accuracy. Furthermore, instead of the moment-based parameterization, we use the original formulation of the Johnson's SU distribution, with location parameter $\xi$, scale parameter $\lambda$, and shape parameters $\gamma$ and $\delta$~\citep{joh:49}. Although these parameters cannot be directly interpreted as moments of the distribution, they allow quantile forecasts to be obtained analytically from the estimated parameters.

For out-of-sample testing, four models, each corresponding to a different hyperparameter set, are retrained daily on the 1,449 most recent observations. DDNN-JSU solves the tabular regression task defined in Eq.~\eqref{eq:tabular}. As in LEAR, the day-of-week variable $\text{DoW}_{d}$ is represented using one-hot encoding. DDNN-JSU uses the same context data as TabPFN-2 and TabPFN-3 but, unlike these models, jointly estimates the distribution parameters for all 24 hourly prices.
For each quantile level, the final forecast is obtained as a trimmed mean of the four quantile forecasts; with four values, this is equivalent to the median. The $\tau$-quantile forecast $\hat{q}\tau$ from each model realization is obtained using the Johnson's SU quantile function~\citep{joh:49}:
\begin{equation}
  \hat{q}_\tau = \hat{\xi} + \hat{\lambda}~\text{sinh}\left(\frac{z_\tau-\hat{\gamma}}{\hat{\delta}}\right),  
\end{equation}
where $z_\tau$ is the $\tau$-quantile of the standard normal distribution, and $\hat{\xi}$, $\hat{\lambda}$, $\hat{\gamma}$, and $\hat{\delta}$ are the estimated distribution parameters returned by the network.

\section{Empirical evidence}
\label{sec:Empirical:Evidence}

\subsection{Statistical error measures and conditional predictive ability}
\label{ssec:Stat:Errors}

The standard metric for evaluating probabilistic forecasts is the Continuous Ranked Probability Score \cite[CRPS;][]{gne:raf:07}: 
\begin{align}
    \operatorname{CRPS}(\hat{F}, x) = \int_{-\infty}^{\infty} \left(\hat{F}(y) - \mathbbm{1}_{\lbrace x \leq y \rbrace} \right)^2 dy,
\end{align}
where $\hat{F}$ is the predictive distribution and $x$ is the observation, e.g., the electricity price $p_{d,h}$. It can be approximated by:
\begin{align}\label{eqn:CRPS:approx}
    \operatorname{CRPS}(\hat{F}, x) \approx \frac{2}{M} \sum_{i=1}^M \operatorname{PS}\left(\hat{q}, x, q_i\right),
\end{align}
where  
\begin{align}
    \operatorname{PS}(\hat{q}, x, q) = \left(\mathbbm{1}_{\lbrace x < \hat{q} \rbrace} - q \right)\left(\hat{q} - x\right)
\end{align}
is the so-called Pinball Score \citep{ber:zie:23,mac:uni:wer:23}, $\hat{q} \equiv \hat{F}^{-1}(q)$ is the quantile forecast for quantile level $q\in(0,1)$, and $\left(q_1, \ldots, q_M\right)$ is an equidistant monotonically increasing dense grid of probabilities. Note that some studies in the EPF literature \citep{mar:nar:wer:zie:23,nit:wer:23,lip:uni:wer:24} report the average Pinball Score across $M=99$ percentiles (quantile levels) and call it CRPS. Clearly, such ``CRPS values'' are half the actual CRPS defined in Eq.~\eqref{eqn:CRPS:approx}.  

The CRPS values below are averages over all hours in the five-year test samples. To complement them, we also compute two standard metrics for point forecasts: the Mean Absolute Error (MAE) and the Root Mean Squared Error (RMSE). As in \cite{mar:nar:wer:zie:23,uni:25:SQRA}, we take the median of the predictive distribution as the point forecast: 
\begin{align}\label{eq:p:median}
    \hat{p}_{d,h} \equiv \hat{p}_{d,h}^{50\%} = \hat{F}^{-1}(0.5).
\end{align}
To formally compare the predictive performance of a foundation model with that of a benchmark, we apply the test of \textit{conditional predictive ability} \citep[CPA;][]{gia:whi:06}. Following Lago et al.~\cite{lag:mar:sch:wer:21} and Lipiecki et al.~\cite{lip:uni:wer:24}, we use its multivariate variant tailored to the 24-dimensional structure of day-ahead electricity price forecasts. The test aggregates the hourly loss differentials within each day and is applied separately using the daily MAE, MSE, and CRPS. In each case, the one-sided alternative is that the foundation model yields a lower expected loss than the benchmark, either LEAR+CP or DDNN-JSU.

\begin{figure}[tb]
    \centering
    \includegraphics[width=.7\linewidth]{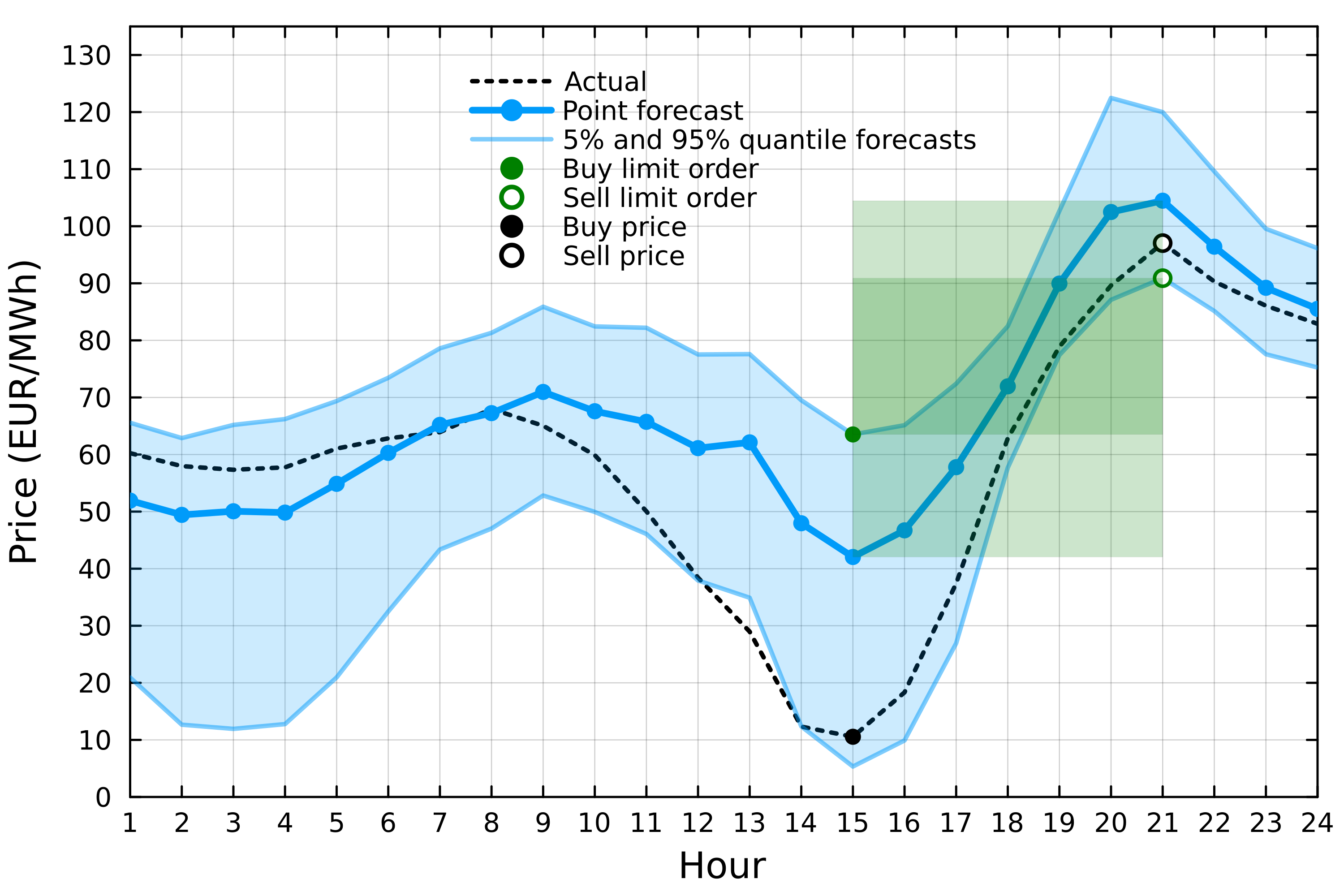}
    \caption{Illustration of the quantile-based trading strategy for TabPFN-3 forecasts, with the QB decision parameter set to $\alpha=90\%$, for a sample day in the German market, 20 April 2024. The height of the light green shaded area represents the predicted profit $\hat{\pi}_d$, see Eq.~\eqref{eq:profit}, associated with buying electricity at hour $h_b=15$ and selling it at hour $h_s=21$. The dark green shaded area represents the corresponding worst-case profit derived from the relevant quantile forecasts, $\hat{p}^{5\%}_{d,h_s}$ and $\hat{p}^{95\%}_{d, h_b}$. Note that on this day the realized spread between the buying and selling prices is greater than expected from the point forecasts. For simplicity, the illustration assumes $\eta=1$.}
    \label{fig:BESS_strategy}
\end{figure}

\subsection{Economic evaluation based on BESS trading}
\label{ssec:Econ:Evaluation}

To evaluate the models in economic terms, we consider two BESS trading strategies similar to those considered by Uniejewski~\cite{uni:25:SQRA}: a \textit{quantile-based} (QB) strategy, with limit-order prices determined by quantile forecasts, and an \textit{unlimited-bid} (UB) strategy; see also the recent review by Hirsch and Ziel~\cite{hir:zie:26}. Both strategies use a 1 MWh battery with a charging and discharging efficiency of $\eta=0.95$, corresponding to a round-trip efficiency of ca. $0.9$. The initial state of charge is 0 MWh. 

In the QB strategy, the trader seeks to generate a profit on day $d$ by buying electricity at hour $h_b$, charging the battery, and subsequently discharging it and selling the electricity at hour $h_s>h_b$; see Figure~\ref{fig:BESS_strategy}. In contrast to the strategy of Uniejewski~\cite{uni:25:SQRA}, we assume that the trader submits a so-called \textit{loop bid}, also referred to as a \textit{loop block} at EPEX SPOT; see \url{https://www.epexspot.com/en/tradingproducts}. The loop bid consists of two coupled orders that are either both executed or both rejected. Consequently, there is no need to settle a partially executed pair of orders in the intraday or balancing market \citep{uni:wer:21}, to trade using only a part of the battery capacity \citep{mac:ser:uni:24,uni:25:SQRA,leb:das:pap:sch:26}, or to replace the limit orders with unlimited bids \citep{mar:nar:wer:zie:23,oco:col:pre:vis:25}.

The trader selects the buying and selling hours, $h_b$ and $h_s$, by maximizing the predicted daily profit:
\begin{align}
    \hat{\pi}_d = \eta\hat{p}_{d,h_s} - \tfrac{1}{\eta}\hat{p}_{d, h_b} - C,
\label{eq:profit}
\end{align}
where $C$ denotes the round-trip operating cost of the battery. This cost accounts for investment expenditure, fixed and variable operation and maintenance costs, and battery degradation. Following the estimate of Lindberg et al.~\cite{lin:zhu:wid:24}, we set $C=25$ EUR. As in Maciejowska et al.~\cite{mac:lip:uni:26}, the same cost $C$ is deducted when calculating the realized profit whenever a trade is executed.

The trader commits to a trade only if $\hat{\pi}_d > 0$ and sets the limit-order prices according to the parameter $\alpha$: the buy-order limit is set to $\hat{p}^{1-\tau}_{d,h_b}$, and the sell-order limit to $\hat{p}^{\tau}_{d,h_s}$, where $\tau=(1-\alpha)/2$. We consider five values of $\alpha \in \{90\%,80\%,\ldots,50\%\}$. 
A higher value of $\alpha$ corresponds to more extreme quantiles and hence less restrictive limit prices, increasing the probability that the loop bid is accepted, see Figure~\ref{fig:BESS_strategy}. We therefore interpret $\alpha$ as a parameter controlling the risk appetite of the trader.
In the unlimited-bid strategy the trader simply submits market orders, that is, buy and sell orders without price limits, whenever the predicted daily profit is positive, $\hat{\pi}_d > 0$.

For comparison, we also consider an \textit{Oracle} (or \textit{Crystal Ball}) strategy, which assumes perfect foresight of day-ahead prices and therefore represents the maximum profit attainable in each market over the test period under the same battery constraints and operating costs.

\begin{table*}[tbp]
\caption{Forecast evaluation for the German power market over the 2021-2025 test period in terms of statistical metrics (MAE, RMSE, and CRPS; see Sec.~\ref{ssec:Stat:Errors}) and economic measures (see Sec.~\ref{ssec:Econ:Evaluation}). For the latter, we report total profits (in EUR) from the 1 MWh battery trading strategy for five levels of $\alpha=90\%, \ldots, 50\%$, as well as for the unlimited-bid (UB) and Oracle scenarios. Cell coloring (\textbf{\dgreen{green}} $\rightarrow$ good, \textbf{\dred{red}} $\rightarrow$ bad) is applied independently for each column. $^*$, $^{**}$, and $^{***}$ indicate that the foundation model significantly outperforms the DDNN-JSU benchmark according to the one-sided multivariate CPA test at the 10\%, 5\%, and 1\% levels, respectively; see Sec.~\ref{ssec:Stat:Errors} for details.}
\vspace{0.1cm}
\label{tab:DE}
\centering
\footnotesize
\begin{tabular}{lccccccccc}
\toprule
& & & &  \multicolumn{6}{c}{\textbf{Total profit}} \\
\multirow{-2}{*}{\textbf{Model}} & \multirow{-2}{*}{\textbf{MAE}}         & \multirow{-2}{*}{\textbf{RMSE}}        & \multirow{-2}{*}{\textbf{CRPS}} & \textbf{UB} & \textbf{90\%} & \textbf{80\%} & \textbf{70\%} & \textbf{60\%} & \textbf{50\%} \\
\midrule
Chronos-2            & \cellcolor[HTML]{D8EDE0}16.13 & \cellcolor[HTML]{C9E7D3}27.00 & \cellcolor[HTML]{D6ECDE}11.80 & \cellcolor[HTML]{B9E1C5}129,689 & \cellcolor[HTML]{8CCF9F}122,257 & \cellcolor[HTML]{89CE9C}111,972 & \cellcolor[HTML]{82CB96}103,184 & \cellcolor[HTML]{7BC890}93,867 & \cellcolor[HTML]{6AC181}85,528 \\
Chronos-2-synth      & \cellcolor[HTML]{CCE8D5}15.77 & \cellcolor[HTML]{CAE8D4}27.07 & \cellcolor[HTML]{CAE7D4}11.53 & \cellcolor[HTML]{A6DAB5}130,337 & \cellcolor[HTML]{AADBB8}119,634 & \cellcolor[HTML]{AADBB9}109,594 & \cellcolor[HTML]{99D4AA}100,990 & \cellcolor[HTML]{A1D7B0}90,680 & \cellcolor[HTML]{82CB96}83,188 \\
Chronos-2-small      & \cellcolor[HTML]{FCEDEF}17.39 & \cellcolor[HTML]{ECF5F1}28.50 & \cellcolor[HTML]{FAFBFD}12.65 & \cellcolor[HTML]{F2F8F7}127,636 & \cellcolor[HTML]{D7EDDF}115,581 & \cellcolor[HTML]{FBEFF2}103,253 & \cellcolor[HTML]{E4F3EA}94,046  & \cellcolor[HTML]{FAB3B6}80,299 & \cellcolor[HTML]{F7FAFB}71,988 \\
Moirai-2             & \cellcolor[HTML]{F8696B}21.84 & \cellcolor[HTML]{F8696B}35.79 & \cellcolor[HTML]{F8696B}16.20 & \cellcolor[HTML]{F8696B}121,823 & \cellcolor[HTML]{D4ECDC}115,909 & \cellcolor[HTML]{E8F4EE}105,160 & \cellcolor[HTML]{D7EDDF}95,257  & \cellcolor[HTML]{FBF1F4}82,566 & \cellcolor[HTML]{F4F9F8}72,350 \\
TimesFM-2.5          & \cellcolor[HTML]{F8696B}18.79 & \cellcolor[HTML]{F8696B}31.32 & \cellcolor[HTML]{F8696B}14.26 & \cellcolor[HTML]{F8696B}124,305 & \cellcolor[HTML]{F8696B}98,639  & \cellcolor[HTML]{F8696B}98,639  & \cellcolor[HTML]{F8696B}77,542  & \cellcolor[HTML]{F8696B}77,542 & \cellcolor[HTML]{F8696B}56,884 \\
TabPFN-2             & \cellcolor[HTML]{63BE7B}12.62$^{***}$ & \cellcolor[HTML]{64BE7C}22.65$^{***}$ & \cellcolor[HTML]{74C489}9.55$^{***}$  & \cellcolor[HTML]{80CA94}131,708 & \cellcolor[HTML]{A2D8B2}120,312 & \cellcolor[HTML]{BAE2C6}108,451 & \cellcolor[HTML]{A1D7B0}100,315 & \cellcolor[HTML]{C0E4CB}88,034 & \cellcolor[HTML]{B5DFC2}78,357 \\
TabPFN-3             & \cellcolor[HTML]{63BE7B}12.59$^{***}$ & \cellcolor[HTML]{63BE7B}22.59$^{***}$ & \cellcolor[HTML]{63BE7B}9.15$^{***}$  & \cellcolor[HTML]{7AC88F}131,919 & \cellcolor[HTML]{9AD5AB}121,026 & \cellcolor[HTML]{B6E0C3}108,730 & \cellcolor[HTML]{AFDDBC}99,013  & \cellcolor[HTML]{9FD7AF}90,787 & \cellcolor[HTML]{9CD5AC}80,768 \\
TabPFN-TS-3          & \cellcolor[HTML]{6AC081}12.81$^{***}$ & \cellcolor[HTML]{66BF7D}22.74$^{***}$ & \cellcolor[HTML]{6AC181}9.33$^{***}$  & \cellcolor[HTML]{63BE7B}132,716 & \cellcolor[HTML]{63BE7B}125,903 & \cellcolor[HTML]{63BE7B}114,705 & \cellcolor[HTML]{63BE7B}105,989 & \cellcolor[HTML]{6FC385}94,907 & \cellcolor[HTML]{6BC182}85,416 \\
Mitra+CP             & \cellcolor[HTML]{C9E7D3}15.69 & \cellcolor[HTML]{AADBB9}25.69 & \cellcolor[HTML]{E2F1E8}12.08 & \cellcolor[HTML]{FBD9DB}126,564 & \cellcolor[HTML]{F8696B}107,863 & \cellcolor[HTML]{F8696B}92,650  & \cellcolor[HTML]{F8696B}79,618  & \cellcolor[HTML]{F8696B}70,058 & \cellcolor[HTML]{F8696B}61,019 \\
\hline 
LEAR+CP              & \cellcolor[HTML]{D1EADA}15.93 & \cellcolor[HTML]{C1E4CC}26.67 & \cellcolor[HTML]{E8F4EE}12.23 & \cellcolor[HTML]{D0EBD9}128,843 & \cellcolor[HTML]{DDF0E5}115,036 & \cellcolor[HTML]{FCFCFF}103,744 & \cellcolor[HTML]{F4F9F8}92,585  & \cellcolor[HTML]{F6FAFA}83,471 & \cellcolor[HTML]{DFF1E6}74,324 \\
DDNN-JSU             & \cellcolor[HTML]{A4D8B3}14.58 & \cellcolor[HTML]{9AD4AA}24.96 & \cellcolor[HTML]{A5D9B4}10.70 & \cellcolor[HTML]{A4D8B3}130,434 & \cellcolor[HTML]{86CD99}122,825 & \cellcolor[HTML]{6BC182}114,197 & \cellcolor[HTML]{73C589}104,520 & \cellcolor[HTML]{63BE7B}95,842 & \cellcolor[HTML]{63BE7B}86,123 \\
\midrule
Oracle & 0 & 0 & 0 & 143,474 & --- & --- & --- & --- & --- \\
\bottomrule
\end{tabular}
\end{table*}

\begin{table*}[tb]
\caption{Forecast evaluation for the Polish power market over the 2021-2025 test period in terms of statistical metrics (MAE, RMSE, and CRPS; see Sec.~\ref{ssec:Stat:Errors}) and economic measures (see Sec.~\ref{ssec:Econ:Evaluation}). The format is the same as in Table~\ref{tab:DE}.}
\vspace{0.1cm}
\label{tab:PL}
\centering
\footnotesize
\begin{tabular}{lccccccccc}
\toprule
& & & &  \multicolumn{6}{c}{\textbf{Total profit}} \\
\multirow{-2}{*}{\textbf{Model}} & \multirow{-2}{*}{\textbf{MAE}}         & \multirow{-2}{*}{\textbf{RMSE}}        & \multirow{-2}{*}{\textbf{CRPS}} & \textbf{UB} & \textbf{90\%} & \textbf{80\%} & \textbf{70\%} & \textbf{60\%} & \textbf{50\%} \\
\midrule
Chronos-2            & \cellcolor[HTML]{C0E3CB}14.76$^{***}$ & \cellcolor[HTML]{D3EBDC}24.59$^{*}$ & \cellcolor[HTML]{BFE3CA}10.74$^{***}$ & \cellcolor[HTML]{B4DFC1}104,839 & \cellcolor[HTML]{8ACE9D}98,642  & \cellcolor[HTML]{A0D7B0}91,073 & \cellcolor[HTML]{A1D8B1}82,869 & \cellcolor[HTML]{C4E6CF}74,291 & \cellcolor[HTML]{BDE3C9}66,101 \\
Chronos-2-synth      & \cellcolor[HTML]{B6DFC3}14.61$^{***}$ & \cellcolor[HTML]{C5E5CF}24.23$^{*}$ & \cellcolor[HTML]{B4DFC1}10.62$^{***}$ & \cellcolor[HTML]{7CC891}106,119 & \cellcolor[HTML]{87CD9B}98,817  & \cellcolor[HTML]{C7E7D1}89,210 & \cellcolor[HTML]{ABDBB9}82,130 & \cellcolor[HTML]{C4E5CE}74,338 & \cellcolor[HTML]{C9E8D3}65,049 \\
Chronos-2-small      & \cellcolor[HTML]{D5ECDD}15.10 & \cellcolor[HTML]{CFEAD8}24.48 & \cellcolor[HTML]{D4EBDC}11.00 & \cellcolor[HTML]{BCE2C8}104,639 & \cellcolor[HTML]{A3D8B2}96,870  & \cellcolor[HTML]{D9EEE1}88,346 & \cellcolor[HTML]{D2EBDB}79,075 & \cellcolor[HTML]{FBE0E3}69,458 & \cellcolor[HTML]{FBF3F6}60,354 \\
Moirai-2             & \cellcolor[HTML]{F8696B}18.21 & \cellcolor[HTML]{F8696B}29.34 & \cellcolor[HTML]{F8696B}13.37 & \cellcolor[HTML]{F8696B}100,886 & \cellcolor[HTML]{C4E6CF}94,430  & \cellcolor[HTML]{FAB6B9}83,448 & \cellcolor[HTML]{F6FAFA}76,252 & \cellcolor[HTML]{FAB7BA}67,308 & \cellcolor[HTML]{FBD8DB}59,250 \\
TimesFM-2.5          & \cellcolor[HTML]{F8696B}16.82 & \cellcolor[HTML]{F8696B}27.47 & \cellcolor[HTML]{F8696B}12.75 & \cellcolor[HTML]{F8696B}99,631  & \cellcolor[HTML]{F8696B}79,381  & \cellcolor[HTML]{F8696B}79,381 & \cellcolor[HTML]{F8696B}63,806 & \cellcolor[HTML]{F87476}63,806 & \cellcolor[HTML]{F8696B}47,677 \\
TabPFN-2             & \cellcolor[HTML]{70C386}13.47$^{***}$ & \cellcolor[HTML]{88CD9B}22.72$^{***}$ & \cellcolor[HTML]{79C78E}9.90$^{***}$  & \cellcolor[HTML]{8CCF9E}105,757 & \cellcolor[HTML]{9FD7AF}97,107  & \cellcolor[HTML]{B0DEBE}90,295 & \cellcolor[HTML]{B9E1C6}80,997 & \cellcolor[HTML]{D2EBDB}73,493 & \cellcolor[HTML]{B5E0C2}66,770 \\
TabPFN-3             & \cellcolor[HTML]{63BE7B}13.25$^{***}$ & \cellcolor[HTML]{63BE7B}21.81$^{***}$ & \cellcolor[HTML]{63BE7B}9.61$^{***}$  & \cellcolor[HTML]{63BE7B}106,685 & \cellcolor[HTML]{70C386}100,529 & \cellcolor[HTML]{97D3A8}91,496 & \cellcolor[HTML]{91D1A3}84,145 & \cellcolor[HTML]{94D2A6}77,208 & \cellcolor[HTML]{96D3A7}69,392 \\
TabPFN-TS-3          & \cellcolor[HTML]{66BF7D}13.30$^{***}$ & \cellcolor[HTML]{74C489}22.23$^{***}$ & \cellcolor[HTML]{64BE7C}9.64$^{***}$  & \cellcolor[HTML]{7FC993}106,060 & \cellcolor[HTML]{63BE7B}101,408 & \cellcolor[HTML]{63BE7B}93,955 & \cellcolor[HTML]{63BE7B}87,706 & \cellcolor[HTML]{6EC385}79,513 & \cellcolor[HTML]{72C488}72,513 \\
Mitra+CP             & \cellcolor[HTML]{C1E4CC}14.78$^{***}$ & \cellcolor[HTML]{A7D9B6}23.50$^{***}$ & \cellcolor[HTML]{EEF6F3}11.33 & \cellcolor[HTML]{FCFCFF}103,158 & \cellcolor[HTML]{F8696B}87,711  & \cellcolor[HTML]{F8696B}79,838 & \cellcolor[HTML]{F8696B}70,638 & \cellcolor[HTML]{F8696B}61,648 & \cellcolor[HTML]{F8696B}54,682 \\
\hline
LEAR+CP              & \cellcolor[HTML]{C6E6D0}14.86 & \cellcolor[HTML]{C8E7D2}24.31 & \cellcolor[HTML]{F8FAFB}11.44 & \cellcolor[HTML]{FBEFF2}102,967 & \cellcolor[HTML]{FBF2F5}90,225  & \cellcolor[HTML]{F87174}80,254 & \cellcolor[HTML]{F88588}71,644 & \cellcolor[HTML]{F8696B}63,190 & \cellcolor[HTML]{F9989A}56,621 \\
DDNN-JSU             & \cellcolor[HTML]{CCE8D6}14.97 & \cellcolor[HTML]{D8EDE0}24.70 & \cellcolor[HTML]{D0EAD9}10.96 & \cellcolor[HTML]{D2EBDB}104,132 & \cellcolor[HTML]{89CE9C}98,688  & \cellcolor[HTML]{76C68C}93,060 & \cellcolor[HTML]{66C07E}87,488 & \cellcolor[HTML]{63BE7B}80,155 & \cellcolor[HTML]{63BE7B}73,731
\\
\midrule
Oracle & 0 & 0 & 0 & 123,439 & --- & --- & --- & --- & --- \\
\bottomrule
\end{tabular}
\end{table*}

\begin{table*}[tb]
\caption{Forecast evaluation for the Spanish power market over the 2021-2025 test period in terms of statistical metrics (MAE, RMSE, and CRPS; see Sec.~\ref{ssec:Stat:Errors}) and economic measures (see Sec.~\ref{ssec:Econ:Evaluation}). The format is the same as in Tables~\ref{tab:DE} and \ref{tab:PL}.}
\vspace{0.1cm}
\label{tab:ES}
\centering
\footnotesize
\begin{tabular}{lccccccccc}
\toprule
& & & &  \multicolumn{6}{c}{\textbf{Total profit}} \\
\multirow{-2}{*}{\textbf{Model}} & \multirow{-2}{*}{\textbf{MAE}}         & \multirow{-2}{*}{\textbf{RMSE}}        & \multirow{-2}{*}{\textbf{CRPS}} & \textbf{UB} & \textbf{90\%} & \textbf{80\%} & \textbf{70\%} & \textbf{60\%} & \textbf{50\%} \\
\midrule
Chronos-2            & \cellcolor[HTML]{EBF5F0}12.94 & \cellcolor[HTML]{D5ECDD}19.47 & \cellcolor[HTML]{E1F1E7}9.46  & \cellcolor[HTML]{CFEAD9}62,871 & \cellcolor[HTML]{7DC992}60,356 & \cellcolor[HTML]{A7DAB6}55,094 & \cellcolor[HTML]{93D2A5}51,338 & \cellcolor[HTML]{BBE2C7}46,577 & \cellcolor[HTML]{B9E1C5}41,992 \\
Chronos-2-synth      & \cellcolor[HTML]{FCF0F3}13.23 & \cellcolor[HTML]{EFF6F3}20.01 & \cellcolor[HTML]{F4F8F8}9.66  & \cellcolor[HTML]{EAF5F0}62,446 & \cellcolor[HTML]{B1DEBE}58,586 & \cellcolor[HTML]{DFF1E6}53,562 & \cellcolor[HTML]{B5DFC2}49,637 & \cellcolor[HTML]{E9F5EF}44,831 & \cellcolor[HTML]{EBF5F0}39,197 \\
Chronos-2-small      & \cellcolor[HTML]{F97E80}13.78 & \cellcolor[HTML]{FBC2C4}20.55 & \cellcolor[HTML]{FBC2C4}10.06 & \cellcolor[HTML]{F9ABAD}61,300 & \cellcolor[HTML]{E2F2E8}56,890 & \cellcolor[HTML]{F9A5A8}51,541 & \cellcolor[HTML]{E7F4ED}47,112 & \cellcolor[HTML]{F99C9F}41,211 & \cellcolor[HTML]{FBE0E2}37,525 \\
Moirai-2             & \cellcolor[HTML]{F8696B}15.33 & \cellcolor[HTML]{F8696B}23.53 & \cellcolor[HTML]{F8696B}11.33 & \cellcolor[HTML]{F8696B}60,600 & \cellcolor[HTML]{C0E4CB}58,061 & \cellcolor[HTML]{E9F5EF}53,286 & \cellcolor[HTML]{C1E5CD}49,000 & \cellcolor[HTML]{FAD2D5}42,854 & \cellcolor[HTML]{FAFCFE}38,344 \\
TimesFM-2.5          & \cellcolor[HTML]{F8696B}13.88 & \cellcolor[HTML]{F8696B}20.94 & \cellcolor[HTML]{F8696B}10.54 & \cellcolor[HTML]{F8696B}59,706 & \cellcolor[HTML]{F8696B}48,606 & \cellcolor[HTML]{F8696B}48,606 & \cellcolor[HTML]{F8696B}38,278 & \cellcolor[HTML]{F8696B}38,278 & \cellcolor[HTML]{F8696B}29,717 \\
TabPFN-2             & \cellcolor[HTML]{63BE7B}11.01$^{***}$ & \cellcolor[HTML]{63BE7B}17.05$^{***}$ & \cellcolor[HTML]{71C387}8.29$^{***}$  & \cellcolor[HTML]{6BC182}64,477 & \cellcolor[HTML]{8BCF9E}59,873 & \cellcolor[HTML]{94D2A6}55,598 & \cellcolor[HTML]{AADBB8}50,188 & \cellcolor[HTML]{D9EEE1}45,454 & \cellcolor[HTML]{CAE8D4}41,010 \\
TabPFN-3             & \cellcolor[HTML]{6FC285}11.18$^{***}$ & \cellcolor[HTML]{78C68D}17.50$^{***}$ & \cellcolor[HTML]{63BE7B}8.15$^{***}$  & \cellcolor[HTML]{78C78E}64,261 & \cellcolor[HTML]{8DCFA0}59,797 & \cellcolor[HTML]{C5E6D0}54,278 & \cellcolor[HTML]{C1E4CC}49,036 & \cellcolor[HTML]{F1F8F5}44,553 & \cellcolor[HTML]{E3F2EA}39,610 \\
TabPFN-TS-3          & \cellcolor[HTML]{83CB96}11.47$^{***}$ & \cellcolor[HTML]{7EC993}17.64$^{***}$ & \cellcolor[HTML]{76C58B}8.35$^{***}$  & \cellcolor[HTML]{63BE7B}64,595 & \cellcolor[HTML]{63BE7B}61,242 & \cellcolor[HTML]{6EC385}56,637 & \cellcolor[HTML]{8CCF9E}51,716 & \cellcolor[HTML]{ABDCBA}47,183 & \cellcolor[HTML]{B6E0C2}42,150 \\
Mitra+CP             & \cellcolor[HTML]{D6ECDE}12.64$^{*}$ & \cellcolor[HTML]{C2E4CD}19.07 & \cellcolor[HTML]{F4F8F8}9.66  & \cellcolor[HTML]{FBE4E7}61,901 & \cellcolor[HTML]{FCFCFF}55,967 & \cellcolor[HTML]{F8696B}50,677 & \cellcolor[HTML]{F8696B}44,976 & \cellcolor[HTML]{F8696B}39,639 & \cellcolor[HTML]{F8696B}34,589 \\
\hline
LEAR+CP              & \cellcolor[HTML]{FCDCDF}13.33 & \cellcolor[HTML]{D0EAD9}19.37 & \cellcolor[HTML]{FBC2C5}10.06 & \cellcolor[HTML]{DDF0E4}62,661 & \cellcolor[HTML]{D2EBDB}57,416 & \cellcolor[HTML]{F9AFB2}51,683 & \cellcolor[HTML]{E9F5EF}46,982 & \cellcolor[HTML]{FAD6D9}42,974 & \cellcolor[HTML]{FBF7FA}38,098 \\
DDNN-JSU             & \cellcolor[HTML]{D7EDDF}12.66 & \cellcolor[HTML]{C2E4CD}19.07 & \cellcolor[HTML]{D1EADA}9.30  & \cellcolor[HTML]{BCE2C8}63,183 & \cellcolor[HTML]{8BCE9D}59,894 & \cellcolor[HTML]{63BE7B}56,930 & \cellcolor[HTML]{63BE7B}53,749 & \cellcolor[HTML]{63BE7B}49,917 & \cellcolor[HTML]{63BE7B}46,714
\\
\midrule
Oracle & 0 & 0 & 0 & 77,558 & --- & --- & --- & --- & --- \\
\bottomrule
\end{tabular}
\end{table*}

\subsection{Results}
\label{ssec:Results}

\subsubsection{Forecast accuracy}

Tables \ref{tab:DE}-\ref{tab:ES} summarize the statistical and economic performance of the benchmark and foundation models in the German, Polish, and Spanish power markets over the 2021-2025 test period. Across all three markets, the strongest statistical performance is delivered by the TabPFN family. In Germany, TabPFN-3 achieves the lowest MAE, RMSE, and CRPS, with values of 12.59, 22.59, and 9.15, respectively. TabPFN-2 and TabPFN-TS-3 perform only slightly worse. A similar pattern is observed in Poland. In Spain, TabPFN-2 achieves the lowest MAE and RMSE, whereas TabPFN-3 produces the lowest CRPS. 

The gains in probabilistic accuracy are substantial. Relative to DDNN-JSU, the stronger of the two EPF-specific benchmarks in terms of the CRPS, TabPFN-3 reduces the CRPS from 10.70 to 9.15 in Germany, from 10.96 to 9.61 in Poland, and from 9.30 to 8.15 in Spain. These reductions correspond to approximately 14.5\%, 12.3\%, and 12.4\%, respectively. The CPA tests confirm that the improvements achieved by the TabPFN models are statistically significant. All three TabPFN variants outperform DDNN-JSU at the 1\% significance level in terms of daily MAE, RMSE, and CRPS in all three markets. The same conclusion holds when LEAR+CP is used as the benchmark; these additional test results are not reported in Tables \ref{tab:DE}-\ref{tab:ES}.

The remaining FMs perform less consistently. The Chronos-2 variants are particularly competitive in Poland, where Chronos-2 and Chronos-2-synth significantly outperform DDNN-JSU across all statistical measures. However, these improvements do not extend to Germany or Spain. Chronos-2-small also yields higher sample errors than the two larger Chronos variants in all three markets.
Moirai-2 and TimesFM-2.5 are consistently among the weakest models and do not significantly outperform DDNN-JSU under any of the considered statistical measures. Mitra+CP performs relatively well in Poland, where it significantly improves on DDNN-JSU in terms of the MAE and RMSE, but not the CRPS. Overall, the results do not support the view that general-purpose time series FMs systematically outperform models designed specifically for EPF.

\subsubsection{Profits from BESS arbitrage}

The economic rankings are less stable and depend strongly on the value of $\alpha$, the decision parameter of the quantile-based strategies. For Germany and Poland, TabPFN-TS-3 generates the highest total profit for the three least risk-averse levels, i.e., $\alpha=90$-$70\%$. For the levels of $60\%$ and $50\%$, DDNN-JSU dominates the economic evaluation in these markets. For Spain, DDNN-JSU performs best for all but the highest level, $\alpha=90\%$. Overall, the QB results reveal a clear transition with the risk level. TabPFN variants tend to perform best under the more risky strategies, whereas DDNN-JSU becomes increasingly competitive as the strategy becomes more risk-averse.

In the unlimited-bid scenario, the highest total profits are obtained by TabPFN-TS-3 in Germany and Spain and by TabPFN-3 in Poland. However, the differences between the leading models are small. For example, in Poland, the unlimited-bid profits of TabPFN-3 and TabPFN-TS-3 are 106,685 EUR and 106,060 EUR, respectively, while in Spain the profits generated by the three TabPFN variants differ by less than 350 EUR. Such small differences, also for some QB configurations, suggest that the precise ranking of closely related models should be interpreted cautiously.

Tables \ref{tab:DE}-\ref{tab:ES} also report results for the \textit{Oracle} strategy, which assumes perfect foresight; hence the zero values in the MAE, RMSE, and CRPS columns. It represents the maximum profit attainable in each market over the test period.
The best UB strategies capture approximately 92.5\%, 86.4\%, and 83.3\% of Oracle profits in Germany, Poland, and Spain, respectively. Thus, although the model-based strategies capture a large share of the profit available under perfect foresight, a non-negligible economic performance gap remains.

\begin{figure}[tb]
    \centering
    \includegraphics[width=\linewidth]{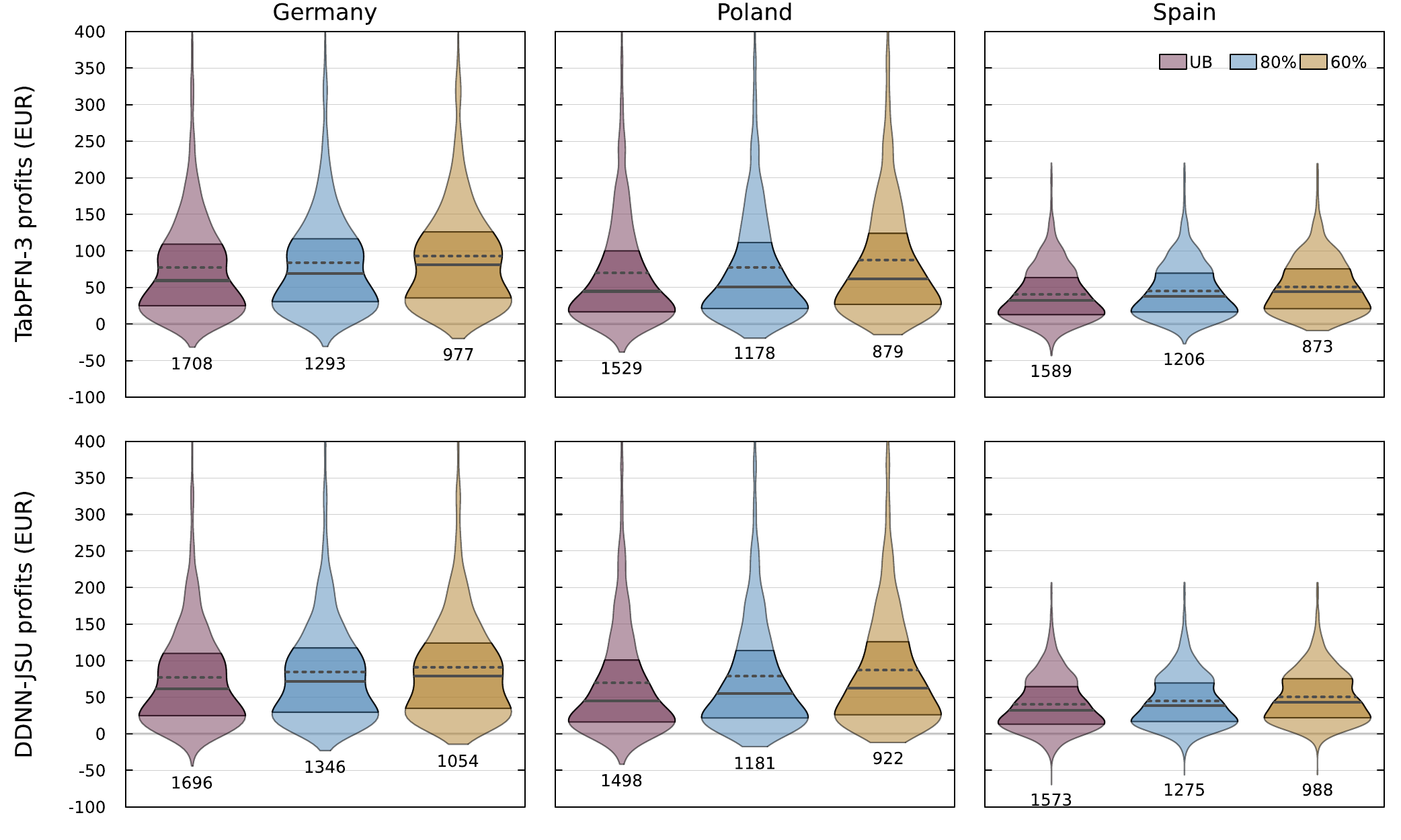}
    \caption{Violin plots of daily profits from settled transactions for the TabPFN-3 (\textit{top row}) and DDNN-JSU (\textit{bottom row}) models in the German, Polish, and Spanish markets (\textit{from left to right}). Each panel shows three trading strategies: the unlimited-bid (UB) strategy and two quantile-based strategies with $\alpha=80\%$ and $\alpha=60\%$. The darker shaded box within each violin indicates the interquartile range, from the 25th to the 75th percentile. The solid line marks the sample median, while the dashed line denotes the sample mean. The number below each violin indicates the number of settled transactions out of the 1,826 trading days for the corresponding model, market, and strategy.}
    \label{fig:profits}
\end{figure}

Figure~\ref{fig:profits} provides further insight into the economic performance of the better-performing benchmark, DDNN-JSU, and the model with the best overall CRPS performance, TabPFN-3, by showing the distributions of daily profits from settled transactions. In all three markets, the distributions are strongly right-skewed, with relatively long upper tails, while some settled transactions result in losses. 
The unlimited-bid strategy produces the largest number of settled transactions, corresponding to around 93\% of the test days in Germany, over 82\% in Poland, and over 86\% in Spain. By comparison, the quantile-based strategies are substantially more selective in all three markets. Across the QB strategies shown in Figure~\ref{fig:profits}, the largest number of settled transactions is 1,346, or about 74\% of the test days, for DDNN-JSU with $\alpha=80\%$ in Germany.
At the same time, the quantile-based strategies generally yield higher mean and median profits per settled transaction, reflecting the avoidance of large losses with the limit bids. Thus, the cumulative profits reported in Tables~\ref{tab:DE}--\ref{tab:ES} reflect two distinct components: the frequency of settled transactions and the distribution of profits conditional on settlement.

The results also show that statistical accuracy does not translate one-to-one into economic performance. The TabPFN models generally rank highly under both evaluation approaches, but the model with the lowest MAE, RMSE, or CRPS does not always generate the highest trading profit. 
DDNN-JSU provides the clearest example: although it does not attain the lowest statistical errors in any of the three markets, it produces the highest profits for $\alpha=60\%$ and $50\%$ in all three markets, as well as for $\alpha=80\%$ and $70\%$ in Spain. Conversely, although TabPFN-3 achieves the best CRPS in every market, TabPFN-TS-3 generates higher profits than TabPFN-3 under every QB configuration considered. However, it is worth noting that the profits generated by the QB strategies considered here are not proper scoring functions \citep{hir:zie:26}. 
Rather, they provide an application-based assessment of the economic value of the forecasts under the specified battery-arbitrage setting.

The comparison across markets further highlights the importance of market-specific conditions. Absolute profits are highest in Germany and lowest in Spain, while the relative ranking of models varies more strongly for trading profits than for statistical measures. TabPFN-3 and TabPFN-TS-3 remain among the statistically strongest models across all three markets, whereas the preferred model under the trading strategy changes with both the market and the value of $\alpha$.

Overall, the results suggest that foundation models can improve electricity price forecasts, but their benefits are model- and application-dependent. The clearest statistical gains are delivered by the tabular TabPFN models, whereas general-purpose time series foundation models are less reliable. At the same time, the differences between statistical and economic rankings confirm that conventional error measures alone are insufficient when forecasts are intended to support trading or storage-operation decisions.

\section{Conclusions}
\label{sec:Conclusions}

This study compares nine variants from five foundation model families with two state-of-the-art electricity price forecasting benchmarks. The models are evaluated over 2021-2025 in Germany, Poland, and Spain using the MAE, RMSE, and CRPS, together with profits from quantile-based and unlimited-bid BESS trading strategies.

Our results show that zero-shot FMs can outperform strong market-specific benchmarks, but that these gains are concentrated in the TabPFN family and do not extend across all model architectures. TabPFN-3 achieves the lowest CRPS in all three markets, as well as the lowest MAE and RMSE in Germany and Poland, while TabPFN-2 performs best according to the latter two measures in Spain. The remaining FMs perform less consistently, although Chronos-2 and Chronos-2-synth yield significant improvements in Poland. Overall, the statistical results do not support the view that FMs can universally replace models designed specifically for EPF.

The economic rankings are more sensitive to the market and trading configuration. TabPFN variants tend to perform best under the unlimited-bid strategy and the riskier quantile-based strategies, whereas DDNN-JSU performs best under the risk-averse strategies.
The distribution of profits from settled transactions further shows that the quantile-based strategies generally trade less frequently but yield higher profits per settled transaction, highlighting the role of risk aversion in determining cumulative profits. Hence, lower statistical errors do not necessarily imply higher cumulative profits. However, these profits are not proper scoring functions; rather, they provide an application-based assessment of the economic value of the forecasts under the specified battery-arbitrage setting.

The usefulness of FMs depends on model architecture, the treatment of exogenous variables, the representation of temporal structure, and the  decision problem. Our analysis is limited to three markets, one battery specification, and one family of trading strategies. The analysis also cannot fully rule out test-period contamination for FMs pretrained partly on empirical time series, although this concern does not apply to models pretrained exclusively on synthetic data, such as the TabPFN variants and Chronos-2-synth. Future studies should consider additional markets, alternative BESS settings, and broader measures of economic performance. Ensembling techniques and probabilistic  FMs should therefore be viewed as promising components of EPF systems rather than universal substitutes for carefully designed market-specific models.

\section*{Acknowledgments}

The study was partially supported by the National Science Centre (NCN, Poland) through grant no.\ 2025/57/B/HS4/02413 (to AL) and grant no.\ 2018/30/A/HS4/00444 (to RW).

\begin{footnotesize}
    \bibliographystyle{elsarticle-num}
    \bibliography{epf}
\end{footnotesize}

\end{document}